\documentclass{ngsm2024} 

\usepackage{nicefrac}

\title[Towards a theory of learning dynamics in deep state space models]{Towards a theory of learning dynamics in deep state space models}

\optauthor{%
\Name{Jakub Smékal} \Email{jsmekal@stanford.edu}\\
\addr Department of Applied Physics, Stanford University \\
\addr Wu Tsai Neurosciences Institute, Stanford University
\AND
\Name{Jimmy T.H. Smith} \Email{jsmith14@stanford.edu}\\
\addr  Institute for Computational and Mathematical Engineering, Stanford University \\
\addr Wu Tsai Neurosciences Institute, Stanford University \\
\addr Liquid AI
\AND
\Name{Michael Kleinman} \Email{mkleinman@stanford.edu}\\
\addr Department of Applied Physics, Stanford University \\
\addr Wu Tsai Neurosciences Institute, Stanford University
\AND
\Name{Dan Biderman} \Email{db3236@cumc.columbia.edu}\\
\addr Center for Theoretical Neuroscience, Zuckerman Institute, Columbia University
\AND 
\Name{Scott W. Linderman} \Email{scott.linderman@stanford.edu}\\
\addr Department of Statistics, Stanford University \\
\addr Department of Computer Science, Stanford University \\
\addr Wu Tsai Neurosciences Institute, Stanford University
}

\begin{document}

\maketitle

\begin{abstract}%

State space models (SSMs) have shown remarkable empirical performance on many long sequence modeling tasks, but a theoretical understanding of these models is still lacking.
In this work, we study the learning dynamics of linear SSMs to understand how covariance structure in data, latent state size, and initialization affect the evolution of parameters throughout learning with gradient descent. 
We show that focusing on the learning dynamics in the frequency domain affords analytical solutions under mild assumptions, and we establish a link between one-dimensional SSMs and the dynamics of deep linear feed-forward networks. 
Finally, we analyze how latent state over-parameterization affects convergence time and describe future work in extending our results to the study of deep SSMs with nonlinear connections.
This work is a step toward a theory of learning dynamics in deep state space models.


\end{abstract}


\section{Introduction}

Deep state space models (SSMs) have become a competitive and efficient building block for long-range sequence tasks \cite{gu2022efficiently, gupta2022diagonal, gu2023mamba, smith2023simplified, hasaniliquid, fuhungry, nguyen2022s4nd, goel2022s, smith2024convolutional, parnichkunstate, patro2024simba, voelker2023legendre, orvieto2023resurrecting}.
Despite a growing body of work \cite{jelassi2024repeat, wang2024stablessm, orvieto2024universality}, there is still little theoretical understanding of the learning dynamics of these models. 
Here, we extend the theory of learning dynamics in deep \textit{linear} feed-forward networks~\cite{baldi1989neural, saxe2014exact, Saxe2019-mi} to the case of \textit{linear} SSMs. By studying this setting, we show how learning is affected by the covariance structure in data and model parameterization. As shown by \citet{saxe2014exact}, the insights gained by studying deep linear models can translate to nonlinear models under mild assumptions. Our present contributions are threefold:

\begin{itemize}
    \item We derive analytical solutions for the learning dynamics of a simplified one-layer SSM by analyzing gradient descent on a squared loss in the frequency domain.
    \item Using these solutions, we establish a link to the time-course of learning in deep linear feedforward networks, connecting existing theory of learning dynamics to SSMs \cite{baldi1989neural, saxe2014exact, Saxe2019-mi}. 
    \item We provide analytical solutions describing the role of over-parameterization in convergence time for a linear $N$-dimensional one-layer SSM.
\end{itemize}
This work is an important step toward a holistic understanding of how data covariance structure, latent state size, and  initialization affect the model's learning dynamics, which could lead to significant improvements in designing future deep SSMs and understanding their training behavior. We provide proofs for our main results in Appendix \ref{sec:proofs}.

\section{State space models in the Fourier domain}
\begin{figure}[t]
    \centering
    \includegraphics[width=\textwidth]{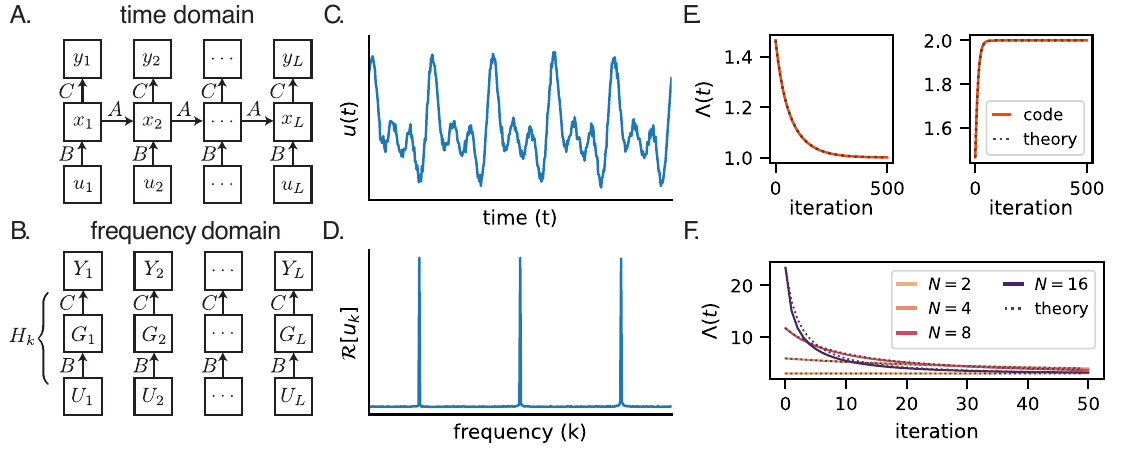}
    \caption{Learning dynamics of SSMs in the frequency domain. \textbf{A.} A linear SSM defined in eq.~\eqref{eq:ssm} unrolled for a length $L$ sequence. \textbf{B.} Applying the discrete Fourier transform, the SSM is fully described by its frequency response $H_k$, transforming a recurrence in the time-domain to modulated scalar multiplication in the frequency domain. \textbf{C.} An example input signal in the time-domain. \textbf{D.} The discrete Fourier transform of the input signal in the frequency domain. \textbf{E.} Even under strong assumptions, the analytical learning dynamics from eq.~\eqref{eq:invtimecourse} approximate the empirical evolution of the SSM for simple input-output modes. Each subplot shows the evolution of the frequency response for individual input-output pairs. \textbf{F.} Extending the theory to $N$-dimensional one-layer SSMs, we show how over-parameterization in the latent state can lead to faster convergence. Full lines denote trajectories arising from automatic differentiation, dashed trajectories are obtained from numerical simulations of the analytical solution to the learning dynamics.}
    \label{fig:ssm_fft_diagram_2}
\end{figure}
\noindent We consider linear time-invariant systems. We begin with a single-input, single-output discrete-time state space model,

\begin{align}
    x_t =Ax_{t-1}+Bu_t, \quad  \quad \quad  \quad
    y_t = Cx_t \label{eq:ssm}
\end{align}
where~$u_t\in\mathbb{R}$ represents the input at time~$t$, $x_t\in\mathbb{R}^N$ represents the latent state, and~$y_t\in\mathbb{R}$ is the output signal. The SSM is  parameterized by the state-transition matrix $A\in\mathbb{R}^{N\times N}$, input vector~$B\in\mathbb{R}^{N\times 1}$, and output vector~$C\in\mathbb{R}^{1\times N}$.  

Gradient descent dynamics in the time domain are complicated by the temporal recurrence, which would typically entail backpropagation through time. 
In the frequency domain, however, the SSM admits a much simpler representation as an element-wise multiplication. 
Moreover, the learning dynamics in the time and frequency domains are equivalent up to a multiplicative constant. 
These properties are summarized in the following propositions. Figure~\ref{fig:ssm_fft_diagram_2}A-D shows an illustration depicting the transformation from a latent state recurrence to scalar multiplication under the discrete Fourier transform.

\begin{proposition} 
\label{prop:dft}
Let $U_k \in \mathbb{C}$ and $Y_k \in \mathbb{C}$ for $k=1,\ldots,L$ denote the discrete Fourier transform~(DFT) of the inputs $u_{1:T}$ and outputs $y_{1:T}$, respectively.
For diagonal dynamics matrices~$A=\mathrm{diag}(a_1,\ldots, a_N)$ with $|a_i|<1$ for all $i=1,\ldots,N$ to ensure stability, the SSM in eq.~\eqref{eq:ssm} is fully described by its frequency response, $Y_k = H_k U_k$, where $H_k\in\mathbb{C}$ is given by,
\begin{align}
H_k 
&= C G_k B, 
&
G_k &= (I - e^{-j\frac{2\pi k}{L}}A)^{-1}.
\label{eq:freq-response}
\end{align}
\end{proposition}

\begin{proposition} \label{prop:graddescent}
The learning dynamics of gradient descent on a squared loss in the frequency domain, $\tilde{\mathcal{L}}=\sum_{k=1}^L|\hat{Y}_k-Y_k|_2^2$, are related to dynamics in the time domain by a proportionality constant given by the length of the sequence.
\end{proposition}

\section{Simplified learning dynamics} \label{sec:simplified}

We derive the continuous-time learning dynamics for a one-layer SSM in the frequency domain. For the full derivation and extension to a two and $K$-layer case, see Appendix \ref{sec:full_derivs}. The continuous-time dynamics equations for parameters $\theta \in \{A, B, C\}$ under a squared error loss function are given by:

\begin{align}
\tau \frac{d \theta}{dt} &= \sum_k\left(Y_kU_k^* - H_k U_k U_k^*\right) \left(\frac{\partial H_k}{\partial \theta}\right)^*. \label{eq:full-dynamics}
\end{align}
These continuous-time ODEs represent a general form of the learning dynamics of a one-layer linear SSM, but they are too complex to offer an intuitive understanding of how the model converges under gradient descent. 
The challenge is that the dynamics for $A$, $B$, and $C$ are nonlinearly coupled, since $H_k$ depends on all three parameters.
To gain some intuition for these dynamics, let us consider how gradient descent behaves under some simplifying assumptions.

First, consider the case of a one-layer, one-dimensional (${N=1}$) SSM with $A \in \mathbb{R}$ fixed, so that the only learnable parameters are $B, C \in \mathbb{R}$.
(In practice, previous work has found reasonable results even when $A$ is fixed~\cite{gu2021combining}.)
Under these simplifying assumptions, the dynamics for $B$ and $C$ are,
\begin{align}
    \tau \frac{d C}{d t} &= (\sigma - CB \eta) B  &
    \tau \frac{d B}{d t} &= (\sigma - CB \eta) C
    \label{eq:cb_dynamics}
\end{align}
where $\sigma = \sum_k Y_k G_k^* U_k^*$ and $\eta = \sum_k U_k U_k^* G_k G_k^*$ are sufficient statistics that summarize the input-output covariances in the frequency domain. 
Thus, we obtain a two-dimensional nonlinear system in which each coordinate's dynamics are conditionally linear given the other.

Following~\citet{saxe2014exact}, we obtain a closed form solution by further assuming that $C=B$. (A solution for $C \neq B$ can be obtained with a hyperbolic change of coordinates, following~\citet[App. A]{saxe2014exact}.) 
The dynamics of the constrained system are characterized by the dynamics of the product, $\Lambda = CB$. Under the assumption that $C=B$, it follows from eq.~\eqref{eq:cb_dynamics} that,
\begin{align}
    \tau \frac{d\Lambda}{dt}
    &=2\Lambda(\sigma-\Lambda \eta)
    \quad \Rightarrow \quad 
    \Lambda(t) = \frac{\sigma}{\eta} \Bigg[ \frac{e^{2\sigma t/\tau}}{e^{2\sigma t/\tau}-1 + \frac{\sigma /\eta}{ \Lambda_0}}\Bigg]. 
    \label{eq:invtimecourse}
\end{align}
From~\eqref{eq:invtimecourse}, we see that $\Lambda(t)$ converges to its limiting value of $\nicefrac{\sigma}{\eta}$ with a time constant $\nicefrac{\tau}{2 \sigma}$. Intuitively, in this simplified regime, the time constant of learning is inversely proportional to the total input-output covariances in the frequency domain. In other words, stronger input-output covariances will lead to faster convergence. Furthermore, this result suggests that for any given sequence data, the strongest covariances will be learned first.
This recovers a major result from the analysis of two-layer feed-forward neural networks from \citet{saxe2014exact}, establishing a link between SSMs in the frequency domain and feed-forward neural networks under particular simplifying assumptions.

Figure~\ref{fig:ssm_fft_diagram_2}E shows a numerical simulation of eq.~\eqref{eq:invtimecourse} faithfully reproducing the dynamics with automatic differentiation. We now proceed to relax the earlier assumptions on latent state size to consider the dynamics of higher-dimensional models.

\section{Learning dynamics for larger latent state sizes} \label{sec:timecourse}
The number of error minima grows to infinity as we increase the latent state size $N$, making each limiting value sensitive to the parameter initialization. To address this analytically, let us consider a symmetric initialization of all parameters across the latent state dimensions, i.e. $A = aI$, $B = b 1_N$, and $C = c 1_N^\top$ for $a, b, c \in\mathbb{R}$. Now, assuming balance $b=c$ and treating $a$ as fixed, as in Section~\ref{sec:simplified}, we arrive at the dynamics of the product $\Lambda=cb$,
\begin{align}
    \Lambda(t) = \frac{\sigma}{\eta} \Bigg[\frac{e^{\frac{2N\sigma t}{\tau}}}{e^{\frac{2N\sigma t}{\tau}}-N + \frac{\sigma/\eta}{\Lambda_0}}\Bigg]. \label{eq:lambda4}
\end{align}
Compared to the one-dimensional case described in eq.~\eqref{eq:invtimecourse}, eq.~\eqref{eq:lambda4} converges to its fixed point value in $\mathcal{O}\left(\frac{\tau}{N\sigma}\right)$ time. As a result, we see that parameterizing the components of the SSM with more latent state dimensions can speed up learning convergence. Figure~\ref{eq:freq-response}F shows this effect across a sweep of latent state sizes $N$ for a synthetic fixed input-output pair. Furthermore, the learning time is still inversely proportional to the sufficient statistics $\sigma$. Relaxing the assumption of balance between $B$ and $C$, we found regimes where the time-course of learning exhibits an inverse quadratic dependence on the latent state-size. We study these other regimes in Appendix \ref{appendix:deeper}. The assumptions made here and in Section~\ref{sec:simplified} uncover close resemblance of the dynamics of learning in $N$-dimensional one-layer linear SSMs to deep feed-forward linear networks. We expect these results to be of relevance to the study of stacked linear SSM layers with nonlinear connections in a similar way that the deep linear feed-forward network analysis in \citet{saxe2014exact} demonstrated connections to nonlinear feed-forward networks. A remaining major challenge is finding the regimes which yield analytical or approximate explanations of the dynamics of the state-transition matrix $A$ with gradient descent, incorporating all sources of nonlinear interactions in eq.~\eqref{eq:full-dynamics}. We leave this analysis to future work.

\section{Conclusion}
In this work, we derived analytical forms of the dynamics of linear state space models with gradient descent on a squared loss. Under mild assumptions, we found a solution showing how sufficient statistics of the data affect convergence time and linked this result to existing theory of learning dynamics in deep feed-forward neural networks. We then extended the analysis to describe the dynamics in $N$-dimensional linear SSMs, concluding that over-parameterization can lead to faster learning in a constrained regime.

In future work, we plan to extend our analysis to understand the role of data and parameterization in the learning dynamics of multi-layer SSMs, with either linear or nonlinear connections between layers. Following prior work in the theory of deep feed-forward networks, we believe the results obtained here will provide a useful starting point when considering these more complex settings.


\bibliography{main}
\appendix

\section{Proofs}\label{sec:proofs}

\begin{proof}[Proposition~\ref{prop:dft}]
Unrolling $y_t$ in~\eqref{eq:ssm} for all points in the sequence with $x_0=0$, the SSM can be represented by a convolution, given by the impulse response of the linear time-invariant system,
\begin{align}
     y_t &= (h * u)_t = \sum_{i=1}^{t} CA^{t-i}Bu_i.
     \label{eq:conv}
\end{align}
To efficiently compute the outputs of the convolution, \citet{gu2022efficiently} made use of the discrete Fourier transform to compute the outputs in the frequency domain before projecting back to the desired state-space. The mapping between the outputs in the time and frequency domains is afforded by the discrete convolution theorem.
Let $Y_k=\mathcal{F}(y_t)$, $U_k=\mathcal{F}(u_t)$, where $\mathcal{F}$ is the discrete Fourier transform. Then

\begin{align}
Y_k &= \mathcal{F}\left(\sum_{i=1}^{t} CA^{t-i}Bu_i\right) \\
          &= \sum_{t=-\infty}^{\infty} \left(\sum_{i=1}^{t} CA^{t-i}Bu_i\right) e^{-j\frac{2\pi k}{L} t} \\
          &= \sum_{t=i}^{\infty} \sum_{i=1}^{\infty} CA^{t-i}Bu_i e^{-j\frac{2\pi k}{L} t} \\
          &= \sum_{m=0}^{\infty} \sum_{i=1}^{\infty} CA^{m}Bu_i e^{-j\frac{2\pi k}{L} (m+i)}
\end{align}
where we applied the transformation $m=t-i$ and noted that $u_t$ is defined for $t\geq 1$. 
Rearranging terms and using the definition of the DFT, $U_k=\sum_{i=1}^{\infty} u_i e^{-j\frac{2\pi k}{L} i}$,
we obtain,
\begin{align}
Y_k 
&= \sum_{m=0}^{\infty} CA^{m}B e^{-j\frac{2\pi k}{L} m} \sum_{i=1}^{\infty} u_i e^{-j\frac{2\pi k}{L} i} \\
&= C \left( \sum_{m=0}^{\infty} A^{m} e^{-j\frac{2\pi k}{L} m} \right) B U_k 
\end{align}
Assuming $|A e^{-j\frac{2\pi k}{L}}| < 1$, which is a necessary condition for the stability of linear time-invariant systems, we can use the geometric series formula,
\begin{align}
    \sum_{m=0}^{\infty} \left(A e^{-j\frac{2\pi k}{L}}\right)^{m} = \left(I - A e^{-j\frac{2\pi k}{L}}\right)^{-1}.
\end{align}
Substituting this into the expression for $Y_k$,
\begin{align}
Y_k &= C (I - A e^{-j\frac{2\pi k}{L}})^{-1} B U_k = H_k U_k
\end{align}
where we have defined $H_k=C (I - A e^{-j\frac{2\pi k}{L}})^{-1}B$ as in Proposition~\ref{prop:dft}.
\end{proof}

\begin{proof}[Proposition~\ref{prop:graddescent}]
By Parseval's theorem, the value of the loss function in the frequency domain is proportional to its time domain counterpart with the proportionality factor given by the length of the sequence considered. Formally,
\begin{align}
    \mathcal{L} = \sum_{t=1}^L|y_t-\hat{y}_t|_2^2, \quad \mathcal{L} = \frac{1}{L}\tilde{\mathcal{L}} = \frac{1}{L}\sum_{k=1}^L|Y_k-\hat{Y}_k|_2^2
\end{align}
where $\hat{y}_t, \hat{Y}_k$ are the model outputs in the time and frequency domains, respectively.
Likewise, their gradients $d \mathcal{L} / d\theta$ and $d \tilde{\mathcal{L}} / d\theta$ are proportional to one another. 
Thus, minimizing the squared error in the time and frequency domains are equivalent, and the dynamics of gradient descent on these two objectives are the same up to a multiplicative factor.
\end{proof}

\section{Further analysis of $N$-dimensional SSMs}\label{appendix:deeper}
Here we share some preliminary analysis showing different ways of deconstructing the training dynamics of the learnable SSM components. An alternative approach to analyze the $N$-dimensional one-layer SSM is to fix both $A$ and $B$ and only learn $C\in\mathbb{R}^{1\times N}$.
(Following common practice from previous work \cite{gu2021combining}.) As in the main text, to reduce the number of fixed points for $N$ latent parameters, we initialize the components across all dimensions to the same value, reducing the sum in (\ref{eq:freq-response}) to a multiplication by $N$, i.e. $(A_{ii}, B_i, C_i)= (a, b, c)\in\mathbb{R}$ for $i=1,...,N$.

\begin{proposition}
\label{prop:timecourse}
Under simplifying assumptions and keeping all other parameters fixed, the time constant of learning $C$ in an $N$-dimensional one-layer SSM scales as $\mathcal{O}\left(\frac{\tau}{N^2\eta}\right)$ where $\eta$ represents the sufficient statistics of the input-input covariances.
\end{proposition}
Consider the case $L=1$, i.e. a single input-output pair in the frequency domain; we leave the extension to arbitrary $L$ to future work. Then the frequency response $H_k=H_1$ becomes
\begin{align}
    H_1 = \sum_{i=1}^{N}\frac{b_i c_i}{1-a_i}.
\end{align}
Following the simplifications of homogeneity across the latent dimensions and fixing $b_i=1$, we obtain the form
\begin{align}
    H_1 = \frac{Nc}{1-a}
\end{align}
where $N$ is the latent state size. Following the same steps as in Section~\ref{sec:simplified}, we arrive at the ODEs describing the time-evolution of $c$ and $a$, which make up the vector $C$ and diagonal matrix $A$,
\begin{align}
    \tau\frac{dc}{dt} &= \frac{N}{1-a}\left(\sigma - \frac{Nc}{1-a}\eta\right) \label{eq:dcdt}\\
    \tau\frac{da}{dt} &= \frac{Nc}{(1-a)^2}\left(\sigma - \frac{Nc}{1-a}\eta\right). \label{eq:dadt}
\end{align}
Both (\ref{eq:dcdt}) and (\ref{eq:dadt}) are separable ODEs. Treating $A$ as fixed and integrating (\ref{eq:dcdt}), we obtain
\begin{align}
    t = \frac{-\tau(1-a)^2}{N^2\eta}\log{\left(\frac{(a-1)\sigma+Nc_f\eta}{(a-1)\sigma+Nc_0\eta}\right)}. \label{eq:time-c}
\end{align}
Solving for $c_f$, we recover the continuous-time evolution of $c$ with gradient descent,
\begin{align}
    c(t) = \frac{e^{-\frac{t N^2 \eta}{\tau (1 - a)^2}} \left((a - 1) \sigma + N c_0 \eta\right) - (a - 1) \sigma}{N \eta} \label{eq:cf}
\end{align}
where in this case $\sigma=\sum_k Y_kU_k^*$ and $\eta=\sum_k U_kU_k^*$,
showing the dependence on both latent state size as well as the sufficient statistics and the initialization of $A$.
The first observation from eq.~\eqref{eq:cf} is that learning converges in $\mathcal{O}\left(\frac{\tau}{N^2\eta}\right)$ in $N$, i.e. over-parameterization improves learning convergence. Furthermore, unlike in (\ref{eq:invtimecourse}), the order of the time-course of learning $C$ is inversely proportional to the strength of the input covariance $\eta$, i.e. stronger input modes speed up learning $C$. We may also be interested in the time-evolution of $A$ under fixed $B$ and $C$. Integrating (\ref{eq:dadt}), we obtain an expression describing the time-course of learning for an $N$-dimensional diagonal $A$,
\begin{align}
    t &= \frac{\tau}{6Nc\sigma^4}\Biggl[6N^3c^3\eta^3\log{\left(\frac{(a_0-1)\sigma+Nc\eta}{(a_f-1)\sigma+Nc\eta}\right)} \\
    &+\sigma((a_f-1)(-3(a_f-1)Nc\eta\sigma + 2(a_f-1)^2\sigma^2+6N^2c^2\eta^2) \notag \\
    -&(a_0-1)(-3(a_0-1)Nc\eta\sigma + 2(a_0-1)^2\sigma^2+6N^2c^2\eta^2))\Biggl]. \notag
\end{align}

\section{Full Derivation of Learning Dynamics}\label{sec:full_derivs}

Here we provide the full derivation of the learning dynamics found in eq.~\eqref{sec:simplified}. Given a sequence transformed via the Discrete Fourier Transform (DFT) to the frequency domain, with inputs $U_k\in \mathbb{C}$  and outputs $Y_k\in \mathbb{C}$, the full one-layer SSM in the frequency domain is defined at all frequency bins $k$ by the frequency response of the linear time-invariant system,
\begin{align}
    H_k 
&= C G_k B 
&
G_k &= (I - e^{-j\frac{2\pi k}{L}}A)^{-1}.
\end{align}
For a single-input, single-output system with a diagonal dynamics matrix $A=\text{diag}(a_1, \dots, a_N)$, the frequency response can be written as,
\begin{align}
    H_k &= \sum_{i=1}^N c_ig_{ki}b_i
    &
    g_{ki} = (1 - e^{-j\frac{2\pi k}{L}}a_i)^{-1}
\end{align}
where $c_i$, $b_i$ are the vector components of $C$ and $B$, respectively, and where we noted that $G_k$ is a diagonal matrix with $i$ indexing its diagonal entries.

We now consider learning the SSM parameters $A, B, C$ of $H$ that best map $U_k$ to $Y_k$ under the squared error loss,
\begin{align}
    \tilde{\mathcal{L}} = \sum_{k=1}^{L} |Y_{k} - H_k U_{k}|^2.
\end{align}
Now we derive an analytical form of the gradients of the learnable parameters $A, B, C$ of $H$ with respect to the loss $\tilde{\mathcal{L}}$. This amounts to computing the gradients of their subcomponents $a_i, b_i, c_i$,

\begin{align}
\Delta A &= \text{diag}(\Delta a_1, \Delta a_2, \dots, \Delta a_N) \nonumber \\
\Delta B &= (\Delta b_1, \Delta b_2, \dots, \Delta b_N) \\
\Delta C &= (\Delta c_1, \Delta c_2, \dots, \Delta c_N) \nonumber
\end{align}
for an $N$-dimensional model. Consider $\Delta a_1$,

\begin{align}
\Delta a_1 &= -\lambda  \frac{\partial \mathcal{L}}{\partial a_1} \label{deltaA1}\\
                 &= -\lambda  \sum_{k=1}^{L} \frac{\partial}{\partial a_1} |Y_{k} - H_kU_{k}|^2 \nonumber \\
                 &= -2\lambda  \sum_{k=1}^{L}   (Y_{k} - H_kU_{k}) U_{k}^* \left(\frac{\partial H_k}{\partial a_1}\right)^* \nonumber \\
                 &= -2\lambda \sum_{k=1}^{L} (\sigma_k-H_k \eta_k)\left(\frac{\partial H_k}{\partial a_1}\right)^* \nonumber
\end{align}
where $\sigma_k=Y_kU_k^*$, $\eta_k=U_kU_k^*$ and we applied the chain rule for complex-valued components. Next, we expand the $\frac{\partial H_k}{\partial a_1}$ term noting that an analogous form can be found for all $\frac{\partial H_k}{\partial a_i}$,

\begin{align}
    \frac{\partial H_k}{\partial a_1} &= \frac{\partial}{\partial a_1} \sum_{i=1}^N c_i(1 - e^{-j\frac{2\pi k}{L}}a_i)^{-1}b_i \\
    &= \sum_{i=1}^N c_i \frac{\partial}{\partial a_1}(1 - e^{-j\frac{2\pi k}{L}}a_i)^{-1} b_i \\
    &= \frac{c_1 b_1 (\cos{\frac{2k\pi}{L}}-j\sin{\frac{2k\pi}{L}})}{(a_1-\cos{\frac{2k\pi}{L}}+j\sin{\frac{2k\pi}{L}})^2}
\end{align}
and similarly for $b_1$ and $c_1$. Extending this for all $i=1,\dots, N$, we get the following continuous-time dynamics for the learnable parameters $A, B, C$:

\begin{align}
\tau \frac{d A}{dt} &= \tau \text{diag}\left(\frac{d a_1}{dt}, \dots, \frac{d a_N}{dt}\right), & \tau \frac{d a_i}{dt} &= \sum_{k=1}^L\left(\sigma_k - H_k\eta_k\right) \left(\frac{\partial H_k}{\partial a_i}\right)^*, \\
\tau \frac{d B}{dt} &= \tau \left(\frac{d b_1}{dt}, \dots, \frac{d b_N}{dt}\right), & \tau \frac{d b_i}{dt} &= \sum_{k=1}^L\left(\sigma_k - H_k\eta_k\right) \left(\frac{\partial H_k}{\partial b_i}\right)^*, \\
\tau \frac{d C}{dt} &= \tau \left(\frac{d c_1}{dt}, \dots, \frac{d c_N}{dt}\right), & \tau \frac{d c_i}{dt} &= \sum_{k=1}^L\left(\sigma_k - H_k\eta_k\right) \left(\frac{\partial H_k}{\partial c_i}\right)^*,
\end{align}
where
\begin{align}
    \frac{\partial H_k}{\partial b_i} &= c_i g_{ki}, & \frac{\partial H_k}{\partial c_i} &= g_{ki}b_i.
\end{align}
The above can easily be extended to a multi-layer SSM by stacking $H^{(l)}$ layers, where the superscript $l$ denotes the index of an SSM layer. Consider, for example, a two-layer SSM parameterized by $A^{(1)}, B^{(1)}, C^{(1)}, A^{(2)}, B^{(2)}, C^{(2)}$:
\begin{align}
    \tau \frac{d a_i^{(1)}}{dt} &= \sum_{k=1}^L\left(\sigma_k - H_k^{(2)}H_k^{(1)}\eta_k\right) \left(\frac{\partial H_k^{(1)}}{\partial a_i^{(1)}}\right)^*\left(H_k^{(2)}\right)^* \\
    \tau \frac{d a_i^{(2)}}{dt} &= \sum_{k=1}^L\left(\sigma_k - H_k^{(2)}H_k^{(1)}\eta_k\right) \left(H_k^{(1)}\right)^* \left(\frac{\partial H_k^{(2)}}{\partial a_i^{(2)}}\right)^*,
\end{align}
with the rest of the parameters following analogous learning dynamics equations.
Finally, we can derive the training dynamics equations for a stacked $K$-layer SSM, noting that diagonal matrices commute:
\begin{align}
\tau \frac{d a_i^{(1)}}{dt} &= \sum_{k=1}^L\left[\left(\sigma_k - \prod_{l=1}^K H^{(l)}\eta_k\right) \left(\frac{\partial H^{(1)}}{\partial a_i^{(1)}}\right)^*\prod_{l=2}^K \left(H^{(l)}\right)^*\right] \\
&\vdots \notag \\
\tau \frac{d a_i^{(K)}}{dt} &= \sum_{k=1}^L\left[\left(\sigma_k - \prod_{l=1}^K H^{(l)}\eta_k\right) \prod_{l=1}^{K-1} \left(H^{(l)}\right)^*\left(\frac{\partial H^{(K)}}{\partial a_i^{(K)}}\right)^*\right].
\end{align}

\end{document}